\pdfoutput=1

\documentclass[11pt]{article}

\usepackage{acl}

\usepackage{times}
\usepackage{latexsym}

\usepackage[T1]{fontenc}

\usepackage[utf8]{inputenc}

\usepackage{microtype}

\usepackage{inconsolata}

\usepackage{graphicx}
\usepackage{multicol}
\usepackage{multirow}

\usepackage{amsfonts}
\usepackage{bbding}
\usepackage{color,xcolor}
\usepackage{ulem}
\usepackage{microtype}
\usepackage{inconsolata}
\usepackage{booktabs}
\usepackage{wrapfig}
\usepackage{lineno}
\usepackage{longtable}

\title{Plan, Generate and Complicate: Improving Low-resource Dialogue State Tracking via Easy-to-Difficult Zero-shot Data Augmentation}

\author{Ming Gu \and Yan Yang\thanks{Corresponding Author} \\
School of Computer Science and Technology, East China Normal University \\
\texttt{51215901012@stu.ecnu.edu.cn, yanyang@cs.ecnu.edu.cn}}

\begin{document}
\maketitle
\begin{abstract}
Data augmentation methods have been a promising direction to improve the performance of small models for low-resource dialogue state tracking. However, traditional methods rely on pre-defined user goals and neglect the importance of data complexity in this task. In this paper, we propose \textbf{EDZ-DA}, an \textbf{E}asy-to-\textbf{D}ifficult \textbf{Z}ero-shot \textbf{D}ata \textbf{A}ugmentation framework for low-resource dialogue state tracking that utilizes large language models to automatically catch the relationships of different domains and then generate the dialogue data. We also complicate the dialogues based on the domain relation to enhance the model's capability for co-reference slot tracking. Furthermore, we permute slot values to mitigate the influence of output orders and the problem of incomplete value generation. Experimental results illustrate the superiority of our proposed method compared to previous strong data augmentation baselines on MultiWOZ.\footnote{\url{https://github.com/SLEEPWALKERG/EDZ-DA}} 
\end{abstract}

\section{Introduction}
The data scarcity challenge in dialogue state tracking (DST) is significant due to the incessant emergence of new domains in task-oriented dialogue systems (ToDs) and the high costs associated with data annotation. Currently, large language models (LLMs) like ChatGPT have shown promising results in zero-shot DST\cite{chatgpt-analyze}. However, although these models achieve superb performance, they have significant limitations such as closed source, request limitations, and deployment difficulties\cite{ldst}. Therefore, a smaller, fine-tuned model is a more practical and cost-efficient choice for DST. Nevertheless, developing such a powerful small model faces a big challenge in the absence of training data.

Recently, strategies of data augmentation with the help of LLM’s strong capability of instruction following and generation is a promising direction for enhancing a task-specific model. However, how to augment the DST data with LLMs is still under-explored. For training a powerful DST model, the logicality and naturalness of the dialogue are important, as well as the data complexity.
\begin{figure}[t]
	\begin{center}
		\includegraphics[width=\columnwidth]{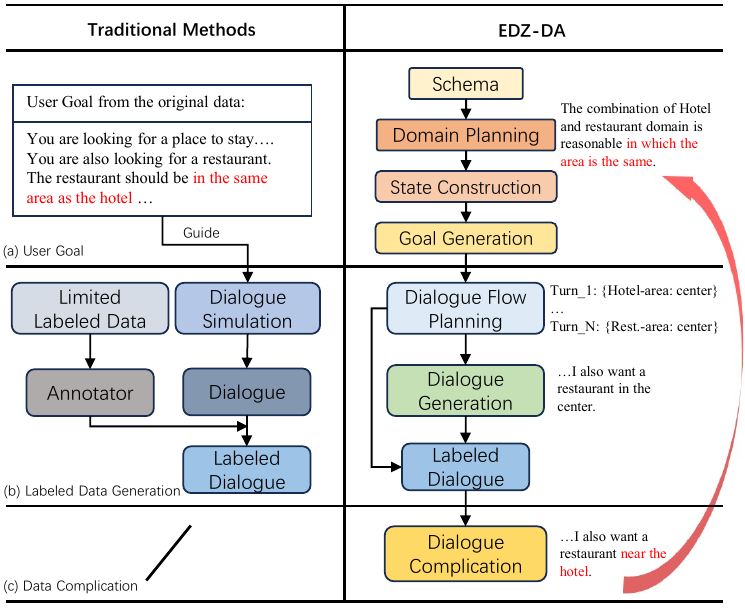}
	\end{center}
	\caption{Differences between traditional methods and our method.}
	\label{point-2-issues}
\end{figure}

We investigate the process of data collection and find that constructing such an annotated dialogue dataset has three major issues: (i) while constructing a dialogue, the user goal is the most important since it guides the whole dialogue construction. Traditional data augmentation methods \cite{neuralwoz, simulated-chats, emnlp2022-unified} directly employ user goals from the original datasets or template-based goals. However, constructing diverse user goals is not that easy. As shown in Figure \ref{point-2-issues}(a), the user goal not only contains all the domain-slots information but also the logical relationship among different domains within a dialogue. We propose to first plan the possible domain combination and then generate the user goal based on the synthetic dialogue state. (ii) annotation accuracy plays an important role in training a DST model. Traditional methods train a model with limited data to annotate the synthesized dialogue. However, due to the limitation of data, the annotator is not satisfying. We propose to plan the dialogue flow first, where the dialogue flow is in the form of the turn state, and then generate the turn utterances based on the turn states. What's more, we propose to first instruct the LLM to generate dialogues with all slot values explicitly appearing in the utterance to alleviate the risk of hallucinations. (iii) complex data is inherently challenging for small models. In multi-domain dialogue state tracking, there are crucial challenges of co-reference, where slot values are sometimes expressed indirectly and should be inferred from the dialogue history. Greater attention should be devoted to this kind of data in order to further enhance the model's ability to handle these challenging samples. However, traditional methods neglect the data complexity problem. We find that the co-reference information like "restaurant-area" shares the same value as "hotel-area" is the direct expression of the domain relationship as shown in Figure \ref{point-2-issues}(c). So, we propose to complicate the dialogue based on the generated logical relationship among domains.

Moreover, for generative information extraction models, the order of the output can exert influence on the model's training\cite{one2set, applying, rl-keyphrase-generation}. For example in value-based DST\cite{svag}, which concatenates multiple slot values in a pre-defined order as the target output during training. Imposing a pre-defined order will result in wrong bias in the training process. And this problem becomes more serious under low-resource settings\cite{ordering-matters}. We propose to permute slot values to mitigate the influence of the output order. Additionally, samples containing several slot values are inherently difficult samples for value generation. The augmentation with permutation can also enhance the model's capability to generate complete slot values within a dialogue turn. 

In this paper, we propose \textbf{EDZ-DA}, an \textbf{E}asy-to-\textbf{D}ifficult \textbf{Z}ero-shot \textbf{D}ata \textbf{A}ugmentation framework for low-resource DST, which leverages the LLM's powerful reasoning ability on dialogue planning and then generate and complicate the dialogue. Specifically, we first propose to automatically catch the logical relationship among different domains with the help of the strong reasoning ability of LLMs and then generate the user goal. Second, we propose to first prompt an LLM to plan the dialogue flow, which contains the turn state annotation, and then generate the corresponding dialogue contents based on the flow, aiming at accurate dialogue generation with annotation matched. Third, we devise to complicate the synthetic dialogues based on the co-reference information to make conversations closer to real scenes and  further improve the state tracker's capability of catching co-reference slots' values. Finally, we also propose to permute slot values to not only mitigate the influence of output orders but also reduce the incomplete generation phenomenon in value generation. Experimental results show that our method outperforms previous data augmentation methods and significantly improves the model's ability for co-reference slots tracking, demonstrating the superiority of our proposed method.

The contributions of this paper are summarized as the following:
\begin{itemize}
	\itemsep0em
	\item We propose EDZ-DA, an effective and generalizable LLM-based easy-to-difficult zero-shot data augmentation framework for low-resource DST.
	\item We propose to plan both the domain relationships and the dialogue flow for natural and accurate labeled dialogue construction. We also devise to complicate dialogues to further enhance model's capability to track co-reference slots.
	\item We propose to permute slot values to mitigate both the influence of the output order and the risk of incomplete generation in the value-based DST model.
	\item Experimental results show that our method achieves new SOTA performance. 
\end{itemize}
\begin{figure*}[htbp]
	\begin{center}
		\includegraphics[width=0.9\textwidth]{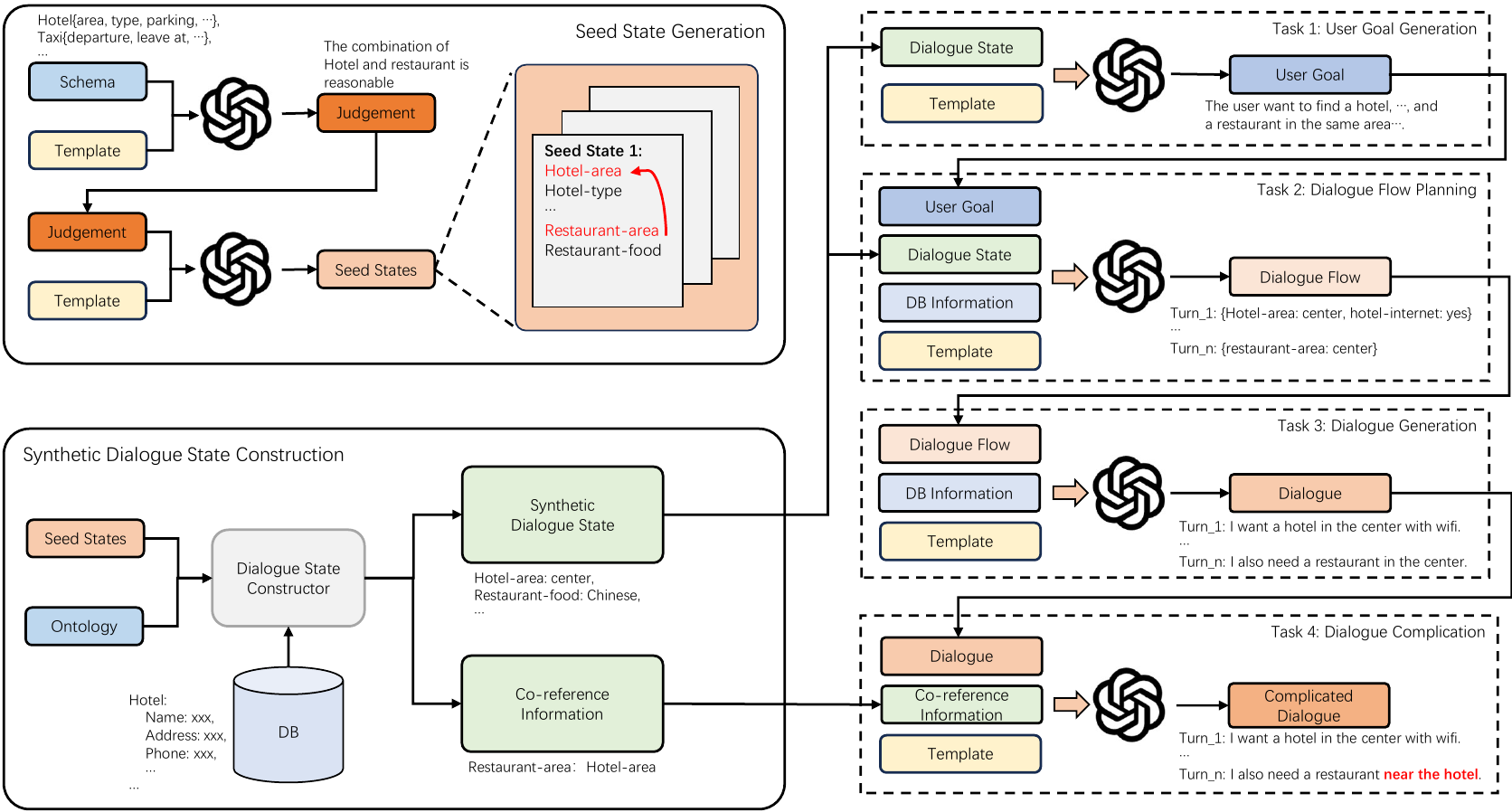}
	\end{center}
	\caption{The overview of our proposed data augmentation method.}
	\label{framework}
\end{figure*}
\section{Methodology}
Figure \ref{framework} illustrates the process of our data augmentation. First, we prompt the LLM to judge whether it is reasonable for different combinations of domains to appear in one dialogue, and then generate the seed state, where the seed state describes the domains and co-reference information within a dialogue. Second, we synthesize diverse dialogue states based on the seed states. Finally, a series of tasks are proposed to generate the labeled dialogue for each synthetic dialogue state and then complicate them based on the co-reference information.

All prompt templates used in our framework are described in appendix \ref{sec:appendix}. Tables \ref{prompt-template-judgement}, \ref{prompt-template-seed-state-generation}, \ref{prompt-template-user-goal}, \ref{prompt-template-dialogue-flow}, \ref{prompt-template-dialogue-generation}, and \ref{prompt-template-dialogue-complication} give an example of our dialogue generation method.
\subsection{Dialogue State Construction}
In this section, we introduce how we construct the synthetic dialogue state.
\subsubsection{Seed State Generation}
For multi-domain task-oriented dialogue, the logicality of the combination of domains is very important, we divide the seed state generation process into two steps: (1) domain judgment and (2) seed state generation. We carefully construct a manual prompt to instruct the LLM to judge whether the combination of domains is logical and reasonable and give some explanation. The MultiWOZ dataset includes five domains, and an analysis of the limited training set reveals that the majority of dialogues encompass one, two, or three domains. Consequently, we have extracted all combinations of two and three domains from the set of five and prompt the LLM to judge the possibility of these combinations of domains within a dialogue. 

Second, we prompt the model to generate several seed states based on the explanation of the judgment and give the co-reference information. The seed state is in forms of a set of domain-slot, value pairs, and the co-reference information is included in it. As shown in figure \ref{framework}, GPT determines that it is possible for hotel and restaurant domains to appear in the same dialogue and "restaurant-area" to share the same value with "hotel-area", which means that the user wants to find a restaurant in the same area as the booked hotel. Since the hallucination problem in LLMs, we carefully construct some rules based on logicality to filter out noisy seed states. For example in Table \ref{prompt-template-seed-state-generation}, "taxi-leaveat: restaurant-book time" in the first seed state is impossible in practice, so we remove it.
\subsubsection{Synthetic Dialogue State Construction}
After obtaining the seed states, we should fill in the blank values in them. For each seed state, we first adopt topological sort to gain the order of domains and then randomly select some places such as restaurants and hotels from the database (DB) according to the domain order. Some of slot values may share the same value with the former domain. So, when encountering these domains, we add constraints while searching the database. The process ends until all values are filled in the seed state. Repeat the aforementioned processes, and we will get several corresponding synthetic dialogue states based on one seed state.
\subsection{Labeled Dialogue Generation}
In this section, we describe how we construct the DST data based on the synthetic dialogue states.
\subsubsection{User Goal Generation}
Although the synthetic dialogue state summarizes the whole dialogue, generating the dialogue based only on the dialogue state suffers from role confusion in that the LLM will assume that the agent already known the requests of the user. Therefore, before generating the dialogue, we first prompt the model to generate the user goal based on the synthetic dialogue state to guide the consequent dialogue flow generation. As shown in Table \ref{prompt-template-user-goal}, we add constraints to ensure the user goal contains only the needs and the named entities like restaurant-names should be recommended by the agent.
\subsubsection{Dialogue Flow Planning}
To correctly generate the dialogue and the corresponding dialogue state annotation, we first prompt the model to plan the flow of a dialogue based on the user goal, the synthetic dialogue state, and the corresponding DB information. The flow consists of a list of \{`description': <description for the user/agent's utterance>, `turn state': <turn state mentioned in the utterance>\} as shown in Table \ref{prompt-template-dialogue-flow}, where the turn state is in the form of a set of domain-slot, value pairs that constraints the content of the current turn. Moreover, we add further information about a certain place from the database and prompt the model to plan some turns to ask for additional information like phone numbers for more natural dialogue generation.
\subsubsection{Easy-to-Difficult Dialogue Generation}
Based on the dialogue flow and the dialogue state, we start to generate the dialogue. The most important thing in dialogue generation is consistency with the dialogue flow because the annotation is the turn label in the dialogue flow. Moreover, we also want the model to generate diverse utterances, especially when encountering co-reference slots. It is difficult to meet the above two needs simultaneously. So, we propose to first generate the dialogue strictly following the dialogue flow and express all slot values in an explicit way. Then we complicate the dialogue turns containing co-reference slots based on co-reference information in the seed state. Table \ref{prompt-template-dialogue-generation} and Table \ref{prompt-template-dialogue-complication} show an example of dialogue generation and dialogue complication, respectively.
\subsection{Slot Value Permutation}
We employ a permutation-based approach to mitigate the influence of sequence order on slot value generation. Table \ref{permutation-example} shows an example. Specifically, we permute the set of slot values within each training example, including every permutation as a distinct training sample. For instance, if the current set of state values is \{"A", "B"\}. the output for the original training example would be "A | B". After permutation, two training samples are generated, one being "A | B" and the other "B | A". This method not only alleviates the impact of output order on the model but also serves as a form of data augmentation. The concurrent generation of multiple state values is one of the inherent challenges in state value generation. The permutation approach significantly amplifies the proportion of such samples within the dataset.
\section{Experiments}
\subsection{Datasets and Metrics}
\textbf{Datasets} We conduct our experiments on the MultiWOZ 2.1 dataset \cite{mwz2-1}. It is a multi-domain task-oriented dialogue dataset which contains 8438 dialogues for training, 1000 dialogues for validating, and 1000 dialogues for testing. Following existing work \cite{trade}, only five domains (restaurant, hotel, attraction, taxi, train) are used in our experiments because the other two domains have very few dialogues and only appear in the training set. We also test our results in MultiWOZ 2.3\cite{mwz2-3} and MultiWOZ 2.4\cite{mwz2-4}. MultiWOZ 2.3 provides the co-reference annotation and MultiWOZ 2.4 is an updated version upon MultiWOZ 2.1, which is the cleanest version of MultiWOZ for testing at the time of writing\footnote{We do not use the validation set of MultiWOZ 2.4 for validating}.

\textbf{Metrics} The standard metric \cite{trade}, joint goal accuracy (JGA) is used in our experiments. This metric compares the whole predicted belief state to the gold one at each dialogue turn. If and only if all the predicted states match the ground truth states exactly for all domains, the prediction is treated as correct. In addition, we adopt co-reference slot accuracy to evaluate the model's capability for tracking co-reference slots.
\subsection{Experimental Settings}
We employ the GPT-4 Turbo model available in OpenAI API\footnote{\url{https://openai.com}} to synthesize all the data. In terms of parameter configuration, a temperature of 0.7 has been set for dialogue generation, aiming at generating more diverse outputs. While for other modules, the temperature has been set to 0. The top-p parameter was uniformly set to 1 for all experiments. 

For the dialogue state tracking model, we use SVAG \cite{svag}, a SOTA small model for low-resource DST, which first generates all slot values in the turn utterances and then generates the corresponding domain-slot type for each generated value. We exclude the self-training strategy in SVAG and directly adopt the experimental settings from \citet{svag}. The base models for both slot value generation and domain-slot generation are T5\cite{t5}, which contains about 770M parameters. Following \citet{trade}, we randomly sampled 1\% and 5\% of the data to simulate the low-resource scenarios with different seeds. We use the same data selection seeds as provided in \cite{svag}, which are 10, 20, and 48.

\subsection{Baseline Models}
We compare our proposed method with several strong baseline data augmentation methods for low-resource DST.
 
\textbf{NeuralWOZ}\cite{neuralwoz} synthesizes annotated dialogues with a collector and a labeler. The collector generates a dialogue by using the given goal instruction and candidate relevant API call results from the KB. The labeler annotated the generated dialogue by reformulating it as a multi-choice problem. The augmented data of NeuralWOZ is publicly available and we sample the same number of dialogues from it for training.

\textbf{Simulated Chats}\cite{simulated-chats} proposes to generate dialogues by simulating the interaction between crowd workers with a user bot and an agent bot. To generate the belief state, they also train a belief state generator. The authors did not provide the augmented data but the code. We reproduce their method and also sample the same number of dialogues from the generated data for training.
\subsection{Main Results}
\begin{table*}[h]
	\centering
	\begin{tabular}{c|c|cc|cc}
		\toprule
		\multirow{2}{*}{\textbf{Augmentation method}} & \textbf{Pre-defined} & \multicolumn{2}{c|}{\textbf{MultiWOZ 2.1}}                    & \multicolumn{2}{c}{\textbf{MultiWOZ 2.4}}                    \\ \cmidrule{3-6} 
		& \textbf{user goal}& \multicolumn{1}{c|}{1}              & 5              & \multicolumn{1}{c|}{1}              & 5              \\
		\midrule
		None        &   -                & \multicolumn{1}{c|}{31.94}  & 43.54  & \multicolumn{1}{c|}{35.53}   & 50.12 \\ \midrule
		NeuralWOZ   &  Manual templates  & \multicolumn{1}{c|}{34.43}  & 43.51  & \multicolumn{1}{c|}{37.64}   & 51.04 \\ \midrule
		Simulated Chats &  Original data & \multicolumn{1}{c|}{-}      & 41.09  & \multicolumn{1}{c|}{-}       & 47.29 \\ \midrule
		Ours w/o P &  -               & \multicolumn{1}{c|}{36.49}  & 43.3   & \multicolumn{1}{c|}{41.79}   & 50.95 \\ \midrule
		Ours & -                      & \multicolumn{1}{c|}{\textbf{37.26}}  & \textbf{44.98} & \multicolumn{1}{c|}{\textbf{43.82}} & \textbf{54.09} \\ 
		\bottomrule
	\end{tabular}
	\caption{Joint goal accuracy of the SVAG models trained on different augmented data. "None" means that only limited original dataset is used for training. P refers to slot value permutation. The average results of 3 runs are reported and best results are marked bold.}
	\label{main-result-ch4}
\end{table*}
\begin{table}[t]
	\centering
	\begin{tabular}{c|cc}
		\toprule
		\multirow{2}{*}{\textbf{Augmentation method}}  & \multicolumn{2}{c}{\textbf{Data ratio}}  \\ \cmidrule{2-3} 
		& \multicolumn{1}{c|}{1}              & 5              \\ 
		\midrule
		None        		& \multicolumn{1}{c|}{30.35}      & 65.63        \\ \midrule
		NeuralWOZ   		& \multicolumn{1}{c|}{35.54}      & 61.87        \\ \midrule
		Simulated Chats     & \multicolumn{1}{c|}{-}          & 54.73        \\ \midrule
		Ours w/o P          & \multicolumn{1}{c|}{64.07}      & 71.08        \\ \midrule
		Ours 				& \multicolumn{1}{c|}{\textbf{64.72}} & \textbf{73.41} 	  \\ 
		\bottomrule
	\end{tabular}
	\caption{Co-reference slot accuracy on MultiWOZ 2.3 of the SVAG models trained on different augmented data.}
	\label{co-ref-ch4}
\end{table}
\begin{table}[t]
	\centering
	\small
	\begin{tabular}{c|c|cc}
		\toprule
		\multirow{2}{*}{\textbf{Model}} & \multirow{2}{*}{\textbf{Param. size}}     & \multicolumn{2}{c}{\textbf{Data ratio}}                      \\ \cmidrule{3-4} 
		&                                  & \multicolumn{1}{c|}{1}              & 5              \\ \midrule
		DS2                & \multirow{6}{*}{\textless{}1B}   & \multicolumn{1}{c|}{36.76}          & 49.89          \\ \cmidrule{1-1} \cmidrule{3-4}
		DS2 + Ours         &    & \multicolumn{1}{c|}{38.99}          & 51.54          \\ \cmidrule{1-1} \cmidrule{3-4}
		SVAG        &    & \multicolumn{1}{c|}{35.53}          & 50.12          \\ \cmidrule{1-1} \cmidrule{3-4}
		SVAG + Ours                         &                                  & \multicolumn{1}{c|}{\textbf{43.82}} & \textbf{54.09} \\
		\midrule
		SM2-3B                & \multirow{3.8}{*}{\textless{}100B} & \multicolumn{1}{c|}{37.59}          & 49.22          \\ \cmidrule{1-1} \cmidrule{3-4} 
		SM2-11B            &  & \multicolumn{1}{c|}{40.03}          & 51.14          \\ \cmidrule{1-1} \cmidrule{3-4} 
		LDST$^*$              &                                  & \multicolumn{1}{c|}{46.77}          & 56.48          \\ \midrule
		IC-DST          & \textgreater{}100B               & \multicolumn{1}{c|}{48.35}          & 55.43          \\ \bottomrule
	\end{tabular}
	\caption{Joint goal accuracy compared with several strong models for low-resource DST on MultiWOZ 2.4. Bolded numbers indicate best performance on models under 1 billion parameters. *: LDST is trained on MultiWOZ 2.2.}
	\label{different-model}
\end{table}
We randomly select 1\% and 5\% data from the training set to simulate the low-resource scenarios with three different seeds to conduct our experiments and we report the averaged JGA score and co-reference slot accuracy over three runs. Table \ref{main-result-ch4} shows the joint goal accuracy of the SVAG model on the MultiWOZ 2.1 and 2.4 test set when subjected to our data augmentation method and other baselines under different data ratio settings. Our method achieves SOTA performance compared to previous augmentation approaches.

Since the permutation of slot values is a general enhancement for generative extraction approaches like SVAG, we also present the results using only the dialogue data generated by our approach ("Ours w/o P"). We observe that our method engenders a more pronounced improvement under the data ratio setting of 1\%, where it surges ahead of the NeuralWOZ augmentation approach by 4.15 in joint goal accuracy on the MultiWOZ 2.4 test set. Moreover, similar performance has been achieved by the two methods under the data ratio setting of 5\%. And under both data ratio settings, EDZ-DA achieves better performance than simulated chats. Different from these two baselines, our approach does not rely on pre-defined user goals from the original dataset or manually constructed goal templates. Instead, our method automatically identifies the relationships among different domains and then generate user goals, providing a more general solution for constructing ToD data. And the better performance reveals both the logicality and accuracy of our proposed planning process and the effectiveness of our proposed labeled dialogue generation method. In particular, we find that Simulated Chats do harm to the model's performance when 5\% data is available. Simulated Chats relies on the fine-tuning process on the limited data. So, the performance of their methods is limited in low-resource scenarios. Our method first plans the dialogue flow which contains the annotation and then generates the dialogue based on it, leading to more accurate labeled data generation.

Table \ref{co-ref-ch4} shows the co-reference slot accuracy by the SVAG model when enhanced with our data augmentation technique and other baselines under different data ratio settings. Our method achieves the highest increase among the two baseline models. Note that our method brings a 200\% improvement in co-reference slot accuracy under the data ratio setting of 1\%, which demonstrates the efficiency of our proposed easy-to-difficult dialogue generation for enhancing the model's capability to track co-reference slots. Under the data ratio setting of both 1\% and 5\%, our method achieves a measurable improvement in co-reference slot accuracy than NeuralWOZ. NeuralWOZ also brings benefits for co-reference slot tracking when only 1\% of original data is available, but the improvement is very limited. For fine-tuning a powerful small model, complex data is very important since these data are even scarcer in extremely low-resource scenarios. It can be observed that both NeuralWOZ and simulated chats engender adverse effects on SVAG when dealing with co-reference slots under the data ratio setting of 5\%, which not only demonstrates the importance of data complexity for DST but also proves that our method can identify logical relations among domains and generate correct complex data to simulate real conversation and further enhance model performance.

Furthermore, we compare two small models enhanced by our method with other strong DST models containing more than 1 billion parameters. Table \ref{different-model} summarizes the results. We conduct experiments on DS2\cite{ds2} using the same data selection seeds provided in the original paper and observe that our augmentation data can further improve its performance under the data ratio settings of 1\% and 5\%, which demonstrates the quality of our annotated dialogue data. Compared to models with more than 1 billion parameters, it can be observed that SVAG enhanced by our augmented data surpasses SM2\cite{sm2} by a margin of 3.79 and 2.95 in JGA under the data ratio settings of 1\% and 5\%, respectively. LDST\cite{ldst} shows better performance than the enhanced SVAG. However, they use MultiWOZ 2.2 for training and evaluate the results on MultiWOZ 2.4. MultiWOZ 2.2 is a cleaner version than MultiWOZ 2.1. Over 17\% of the annotation in MultiWOZ 2.1 is corrected in 2.2. So, it is not comparable. What's more, the enhanced SVAG model achieves competitive performance with IC-DST\cite{ic-dst}. IC-DST is based on CodeX, which contains more than 100 billion parameters. In summary, our proposed data augmentation method can significantly improve small models' performance in low-resource DST, reaching even better performance than the models ten times larger. Furthermore, since SVAG achieves better performance with the augmented data, it will make the consequent self-training more effective and further improve the performance.
\begin{table}[h]
	\small
	\centering
	\begin{tabular}{l|cc|cc}
		\toprule
		\textbf{Aug.} & \multicolumn{2}{c|}{\textbf{MultiWOZ 2.1}} & \multicolumn{2}{c}{\textbf{MultiWOZ 2.4}} \\ \cmidrule{2-5} 
		\textbf{method} & \multicolumn{1}{c|}{1}              & 5              & \multicolumn{1}{c|}{1}              & 5              \\ \midrule
		EDZ-DA        & \multicolumn{1}{c|}{37.26}          & \textbf{44.98} & \multicolumn{1}{c|}{\textbf{43.82}} & \textbf{54.09} \\
		~~~~-DG & \multicolumn{1}{c|}{34.39}          & 43.76          & \multicolumn{1}{c|}{39.11}          & 52.15          \\ 
		~~~~-P  & \multicolumn{1}{c|}{36.49}          & 43.3           & \multicolumn{1}{c|}{41.79}          & 50.95          \\ 
		~~~~-P-Comp.   & \multicolumn{1}{c|}{\textbf{37.45}} & 43.87  & \multicolumn{1}{c|}{43.71}          & 50.89          \\ 
		\bottomrule
	\end{tabular}
	\caption{Ablation study on joint goal accuracy. Comp. refers to data complication and P refers to slot value permutation.}
	\label{ablation-jga}
\end{table}
\begin{table}[t]
	\small
	\centering
	\begin{tabular}{l|cc}
		\toprule
		\multirow{2}{*}{\textbf{Aug. method}}  & \multicolumn{2}{c}{\textbf{Data ratio}}                      \\ \cmidrule{2-3} 
		& \multicolumn{1}{c|}{1}              & 5              \\ \midrule
		EDZ-DA                    & \multicolumn{1}{c|}{\textbf{64.72}} & \textbf{73.41} \\
		~~~~-DG            & \multicolumn{1}{c|}{47.21}          & 70.56          \\ 
		~~~~-P             & \multicolumn{1}{c|}{64.07}          & 71.08          \\ 
		~~~~-P-Comp.       & \multicolumn{1}{c|}{45.65}          & 63.16          \\ 
		\bottomrule
	\end{tabular}
	\caption{Ablation study on co-reference slot accuracy on MultiWOZ 2.3.}
	\label{ablation-coref}
\end{table}

\subsection{Ablation Study}
We conduct an ablation study to identify the contribution of different components from our proposed augmentation method. Table \ref{ablation-jga} shows the joint goal accuracy score tested on MultiWOZ 2.1 \& 2.4 when trained with different versions of our augmented data and Table \ref{ablation-coref} shows the co-reference slot accuracy tested on MultiWOZ 2.3.

First, we eliminate the generated dialogue data by the LLM, denoted as "-DG". We observe that both the joint goal accuracy and co-reference slot accuracy drop a lot under all data ratio settings. Notably, under the data ratio setting of 1\%, removing the generated dialogue data dramatically harms the model performance, leading to a 4.71 decrease in JGA tested on MultiWOZ 2.4 and a 17.51 decrease in co-reference slot accuracy. The results indicate the effectiveness of our proposed method for DST data generation.

Second, we examine the use of slot value permutation. The results without slot value permutation ("-P") show that removing slot value permutation leads to a decrease in both JGA and co-reference slot accuracy under all data ratio settings, demonstrating the effectiveness of slot value permutation. Slot value permutation can not only mitigate the influence of output orders but also reduce the risk of incomplete generation, leading to better dialogue state tracking performance. Notably, as depicted in Table \ref{ablation-coref}, the co-reference slot accuracy decreases more while removing the generated data, compared to the impact of removing slot value permutation, which further proves the significance of our proposed easy-to-difficult dialogue generation method for co-reference slot tracking.

Third, to evaluate the effectiveness of the dialogue complication strategy, we conduct an experiment with only the generated dialogue data without complication ("-P-Comp."). Under the data ratio setting of 1\%, we observe that the joint goal accuracy score is improved a little on the MultiWOZ 2.1 test set when only dialogue data without complication is used. However, as shown in Table \ref{ablation-coref}, the co-reference slot accuracy drops a lot without dialogue complication. Dialogue complication can significantly improve the model's capability to track co-reference slots. Under the data ratio setting of 5\%, the generated augmentation data without dialogue complication even does harm to the co-reference slots training, which further illustrates the data complexity's importance for DST.
\subsection{Case Study}
\begin{table}[t]
	\centering
	\small
	\begin{tabular}{p{0.95\columnwidth}}
		\toprule
		\textbf{Dialogue history:} ... [user] I would like to go to {\color[RGB]{84, 180, 53}{thanh binh}}. [sys] Excellent. For how many would you like a reservation and at which preferred date and time? [user] I would like the reservation to be for 4 people on thursday at {\color[RGB]{84, 180, 53}{18:30}}, please.\\
		\textbf{Current Turn Utterances:} [sys] you're all set! Your reference number is: xxx. [user] I would also like a taxi to get me to {\color[RGB]{84, 180, 53}{the restaurant by that time}}. \\
		\textbf{Without augmentation:} None\\
		\textbf{Augmented with NeuralWOZ:} taxi-destination: thanh binh, restaurant-book time: 18:30 \\
		\textbf{Augmented with ours:} taxi-destination: thanh binh, \textbf{taxi-arriveby: 18:30} \\
		\textbf{Ground truth:} taxi-destination: thanh binh, taxi-arriveby: 18:30 \\
		\midrule
		\textbf{Dialogue history:} [user] Hello, I am looking for a train that arrives by 16:00 and leaves on {\color[RGB]{84, 180, 53}{Monday}}. ...[sys] The tr0796 train leaves cambridge at 05:01 and arrives in broxbourne 06:01. the cost in 17.90 pounds. Is this a good train for you? [user] Yes please. Please book a ticket for 1 person.\\
		\textbf{Current Turn Utterances:} [sys] The booking was successful your reference number is xxx. Is there anything I can help with today? [user] Yes. I would like a restaurant {\color[RGB]{84, 180, 53}{for the same day}}. the name is {\color[RGB]{84, 180, 53}{travellers rest}}. \\
		\textbf{Without augmentation:} restaurant-name: travellers rest, restaurant-book day: Monday \\
		\textbf{Augmented with NeuralWOZ:} restaurant-name: travellers rest \\
		\textbf{Augmented with ours:} restaurant-name: travellers rest, restaurant-book day: Monday \\
		\textbf{Ground truth:} restaurant-name: travellers rest, restaurant-book day: Monday \\
		\bottomrule
	\end{tabular}
	\caption{Example dialogue state outputs from SVAG ehanced by EDZ-DA and NeuralWOZ under the data ratio setting of 1\%.}
	\label{case-ch4}
\end{table}
In this section, we give some example output of the SVAG model enhanced by EDZ-DA and NeuralWOZ. Table \ref{case-ch4} shows two examples. In the first example, The user express that he/she want a taxi to get to the restaurant by the booking time. Both the models enhanced by NeuralWOZ and our method can capture the slot values for the two shared slots during the value generation stage. However, the model enhanced by NeuralWOZ fails to generate the correct "domain-slot" for the time "18:30", indicating that our data augmentation method can better enhance the model's ability to track co-reference slots. Moreover, it also demonstrates the effectiveness of our method in capturing dialogue logic with more accurate annotations.

In the second example, the user want to book "travellers rest" for the same day with the hotel booking. We find that the model enhanced by NeuralWOZ fails to capture the information "restaurant-book day: Monday," while both the models without augmentation and with our data augmentation can accurately capture the information of the shared slot. This indicates that our proposed complexity-aware method can generate complex data with accurate annotations, thereby ensuring that the augmentation does not weaken the model's reasoning ability but enhances it. In contrast, traditional methods like NeuralWOZ do not pay attention to the importance of data complexity and even weaken the model's reasoning ability by adding too much simple data. This is also shown in Table \ref{co-ref-ch4}, where the 1\% original data can only provide limited reasoning ability to the model, while the 5\% data can already provide some reasoning ability for co-reference slots. Adding too much simple data will weaken the model's ability to track difficult data like co-reference slots.
\subsection{Constraint Following Analysis}
We do an additional evaluation of the LLM's adherence to the imposed constraints during data generation. 95.8\% of the generated dialogues are retained, and the rest are deleted because they cannot match the planned dialogue flow in the process of dialogue generation. Furthermore, we sample some generated dialogue goals for further evaluation and find that all generated goals are in the form of the pre-defined form in the prompt. Additionally, we also sample 50 dialogue turns that contain co-reference slots to manually check the correctness of dialogue complication. 96\% of these turns express the co-reference slots implicitly and correctly after the complication process. In summary, the LLM can follow our instructions well to generate correct and natural dialogues.
\section{Related Work}
\subsection{Low-resource DST}
Low-resource dialogue state tracking has received increasing attention in academia and industry. Most previous work have attempted to tackle the challenge in three ways\cite{survey-2022}: (1) cross-domain transfer learning\cite{trade, meta-learning}; (2) cross-task transfer learning\cite{mrc2dst, transfer}; and (3) pre-trained language model adaption\cite{tod-bert, pptod, cins, soloist, simpletod}. Recently, more and more data augmentation based approaches have been proposed for low-resource DST. \citep{neuralwoz} proposed a collector to synthesize dialogues and a labeler to annotate the generated dialogues.\citep{simulated-chats} proposed to generate dialogue data by mimicking the data collection process employed by crowd workers.\citet{emnlp2022-unified} proposed to first pre-train the user simulation model on several publicly available datasets and then tune it on target domains with few-shot data. However, all of these methods rely on the usage of the user goal from the original dataset.
\subsection{Data Augmentation via LLMs}
Recently, more and more studies have tended to prompt LLMs to generate synthetic training data with the purpose of augmenting data in low-resource scenarios. \cite{zeroshotdataaug, augesc, auggpt, ischatgpt} used GPT-3.5 and GPT-4 as the base generative model for data augmentation. \citet{zeroshotdataaug} evaluated the effectiveness of zero-shot prompting for data augmentation under low-resource settings. \citet{auggpt} used GPT to generate paraphrases of existing texts for augmentation. Both studies report better results using LLMs for data augmentation compared to previous SOTA data augmentation approaches. \citet{ischatgpt} further compared different data augmentation methods and revealed that the performance of ChatGPT is highly dependent on the dataset. The more relevant work to ours is \cite{augesc}, which used LLM to generate open-domain dialogue data for emotional support conversation. In this paper, we prompt the LLM to generate DST data, which is more challenging due to the difficulty in domain planning, the demand for accurate annotation, and the co-reference data. 
\section{Conclusions and Future Work}
In this paper, we propose EDZ-DA, an easy-to-difficult zero-shot data augmentation framework for low-resource DST. We reveal three issues in constructing DST data and propose to first determine the logical relationship among domains and generate the user goal with the help of the LLM's strong reasoning ability. In order to enhance the DST model's performance in tracking co-reference slots, we propose to complicate the dialogue content based on the domain relationship. Moreover, we propose to permute slot values to mitigate the influence of output order and the incomplete generation problem. Experimental results on the MultiWOZ dataset illustrate the superiority of EDZ-DA over previous data augmentation approaches for low-resource DST. 

In future work, we will further study how to generate diverse natural dialogue flows.

\section*{Limitations}
In this section, we discuss several limitations of our proposed framework. First, although our generated augmentation data can significantly improve low-resource DST, the naturalness of the dialogue process can be further improved. In practice, there may be situations such as booking failure and re-qualification. Future work can look into studying how to prompt the LLM to plan such dialogues. Second, the prompts in our method are manually constructed. How to explore a more systematic method for prompt engineering leaves future direction for our work. Finally, it could be interesting to investigate the performance of other LLMs such as LLaMA\cite{llama} in this task.

\section*{Ethics Statement}
In our paper, we propose an LLM-based data augmentation method for low-resource DST. We choose GPT-4 for generating all the augmentation data and use T5 as the backbone model of our DST model. We carefully check all outputs in our experiments and we do not observe any ethical issues. Moreover, we conduct our experiments on the MultiWOZ dataset which is a publicly-available benchmark, and in our view, it does not have any attached privacy or ethical issues. In summary, there are no direct ethical concerns in our study.

\section*{Acknowledgements}
We would like to thank the anonymous reviewers for their valuable comments. This research is funded by the Science and Technology Commission of Shanghai Municipality Grant (No. 22511105901).

\normalem
\bibliography{custom}

\appendix

\section{Appendix}
\label{sec:appendix}
\begin{table*}[h]
	\centering
	\begin{tabular}{p{2\columnwidth}}
		\toprule
		\textbf{Prompt: } \\There are five domains that the AI agent supported, and their slots are listed in the following:\\
		1. Hotel: \{area, type, internet, parking, name, book day, price range, star, book stay, book people\}\\
		2. Restaurant: \{area, book day, book people, book time, food, name, price range\}\\
		3. Attraction: \{area, name, type\}\\
		4. Taxi: \{arrive by, departure, destination, leave at\}\\
		5. Train: \{book people, day, departure, destination, leave at\}\\
		6. Flight: \{arrive by, departure, destination, leave at\}\\
		Categorical slots and their possible values:\\
		1. Area: centre, east, south, west, north\\
		2. Internet: yes, no\\
		3. Parking: yes, no\\
		4. Price range: cheap, moderate, expensive\\
		5. Star: 1, 2, 3, 4, 5\\
		6. Day: Monday, Tuesday, Wednesday, Thursday, Friday, Saturday, Sunday\\
		7. Hotel-type: hotel, guest house\\
		8. Book people: 1, 2, 3, 4, 5, 6, 7, 8\\
		9. Book stay: 1, 2, 3, 4, 5, 6, 7, 8\\
		10. Arrive by \& leave at: time in forms of "xx:xx" such as "13:00"\\
		
		I will give you some of these domains. Your task is to determine whether these domains can be fused together in a dialogue reasonably. The order of these domains does not matter, and you can determine the order of these domains yourself. Please output it in a json format like \{"is\_reasonable": <1 for reasonable and 0 the opposite>, "explanation": <give some explanation about your judgement>\}.\\
		
		{\color[RGB]{84, 180, 53}{(Example-1)\# domains: restaurant, taxi}}. \\
		{\color{red}{(Example-2)\# domains: flight, train}} \\
		\midrule
		\textbf{GPT output of example-1: } \\
		\{\\{\color[RGB]{84, 180, 53}{"is\_reasonable": 1}},\\ "explanation": "The domains of restaurant and taxi can be reasonably fused together in a dialogue. A common scenario is when a user wants to book a restaurant and then needs a taxi to get to the restaurant. The user might want to book a table at a specific time and then arrange for a taxi to arrive or depart accordingly. The slots for restaurant bookings (such as area, book day, book time, and book people) can be discussed first, followed by arranging a taxi (with departure, destination, leave at, and arrive by) to ensure the user arrives on time for their reservation."\\\}\\
		\midrule
		\textbf{GPT output of example-2: } \\
		\{\\{\color{red}{"is\_reasonable": 0}},\\ "explanation": "Fusing the flight and train domains in a single dialogue could be confusing and impractical because they both serve the purpose of transporting a person from one location to another, but they are typically used for different types of journeys. Flights are generally used for longer distances or international travel, while trains are often preferred for regional or shorter distance travel. Including both in a single dialogue might suggest an unrealistic scenario or a misunderstanding of the user's travel needs. Additionally, coordinating between the two modes of transport would involve complex logistics, such as timing and transfer points, which are beyond the scope of a simple dialogue system."\\\}\\
		\bottomrule
	\end{tabular}
	\caption{Two examples of domain relation judgment. GPT determines that it is reasonable for the restaurant domain and taxi domain to appear in one dialogue but fusing the flight and train domains is confusing and impractical.}
	\label{prompt-template-judgement}
\end{table*}
\begin{table*}
	\centering
	\begin{tabular}{p{2\columnwidth}}
		\toprule
		\textbf{Prompt:}\\
		So, based on your judgement, please help me generate several dialogue states with different inferrable information. Each dialogue state is in forms of a dictionary where the key is the domain-slot and the value is the corresponding value. The dialogue state summarizes a user’s goal in his/her dialogue with an AI agent. You should determine the number of states yourself based on your judgement and output your answer with a list of dialogue states.
		
		You can determine the values of the categorical slots yourself. Values of other non-categorical slots should be left blank (an empty string), and I will fill them. Note that values should be logical among different domains. What’s more, some the values may be the value of another domain. Please express this information by "domain1-slot": "domain2-slot". But note that you cannot refer to domains that are not mentioned in the current domains. And there are some restrictions in my system:
		
		1) slots {`departure', `destination'} cannot be inferred from slot `area'.\\
		2) slots {`arrive by', `leave at', `book time'} cannot be inferred from slots {`day', `book day'} and vice versa.\\
		3) area can only be inferred from area.\\
		4) train-destination and train-departure cannot be inferred from other domain.\\
		5) inferable slots must be different domains.\\
		6) Please carefully refer to the schema I have given in the former turn. Only legal slots in a certain domain can be inferred.\\
		7) Infer from a blank slot is allowed since I will fill in all the blank slots later.\\
		\midrule
		\textbf{GPT output:} \\
		Based on your instructions, here are several dialogue states with inferable information, formatted as dictionaries. Each state represents a user's goal in a conversation with an AI agent, with some values left blank for you to fill in later:
		
		[\\
		\{"taxi-departure": "", "taxi-destination": "restaurant-area", "taxi-leave at": "restaurant-book time", "restaurant-book time": "", "restaurant-area": "", "restaurant-food": "Italian"\}, \\
		{\color[RGB]{84, 180, 53}{\{"taxi-departure": "", "taxi-destination": "restaurant-name", "taxi-arriveby": "restaurant-book time", "restaurant-name": "", "restaurant-book time": "", "restaurant-pricerange": "moderate"\}}}, \\
		\{"taxi-departure": "hotel-name", "taxi-destination": "restaurant-name", "taxi-arriveby": "", "hotel-name": "", "restaurant-name": "", "restaurant-book time": "taxi-arriveby"\}, \\
		\{"taxi-departure": "hotel-area", "taxi-destination": "", "taxi-leaveat": "", "hotel-area": "", "restaurant-name": "taxi-destination", "restaurant-booktime": "taxi-leaveat"\} \\
		] \\
		
		In these dialogue states, I've adhered to the restrictions you've provided. The values for categorical slots have been determined, and the non-categorical slots have been left blank. The inferable information is based on the schema you mentioned, ensuring that the slots are inferred from different domains and that the restrictions on which slots can be inferred from which are respected.\\
		\bottomrule
	\end{tabular}
	\caption{An example of seed state generation. Words in green are one of the selected seed states.}
	\label{prompt-template-seed-state-generation}
\end{table*}
\begin{table*}
	\centering
	\begin{tabular}{p{2\columnwidth}}
		\toprule
		\textbf{Prompt:}\\
		Given the following dialogue state with various domain-slot pairings that include specific details:
		
		\{\\
		"restaurant-name": "midsummer house restaurant",\\
		"restaurant-book time": "12:30",\\
		"restaurant-area": "centre",\\
		"restaurant-book people": "1",\\
		"restaurant-food": "british",\\
		"restaurant-book day": "wednesday",\\
		"restaurant-pricerange": "expensive",\\
		"taxi-destination": "midsummer house restaurant",\\
		"taxi-arriveby": "12:30",\\
		"taxi-departure": "peterborough train station"\\\}.
		
		Construct the user’s overall goal for the ensuing conversation based solely on the information provided in the dialogue state. The user goal should be as specific as possible while adhering to the following guidelines:
		
		- Do not include any specific names of restaurants, hotels and attractions that are present in the dialogue state. These details are meant to be suggested by the agent during the conversation.\\
		- Do not infer or add any information beyond what is present in the dialogue state. Use only the details provided in the slots.\\
		- The order in which the domain-slot pairings are presented is not indicative of importance or the sequence of addressing them in the dialogue. The language model should determine the logical flow and prioritization independently.\\
		- Ensure that the user goal lays a foundation for a focused yet dynamic interaction, where the agent has the opportunity to make recommendations and the user can respond or confirm these suggestions.\\
		
		The objective is to clarify what the user intends to achieve or learn during the interaction without presuming or preempting the agent's role in guiding the conversation.
		
		Example of a properly constructed user goal:
		
		"User aims to find [general need based on dialogue state], with preferences matching [specific preferences], and is looking for recommendations on [broad categories without naming specific entities]."
		
		Please proceed by formulating a user goal that aligns with these criteria.\\
		\midrule
		\textbf{GPT output:} \\
		The user aims to book a table for dining at a high-end restaurant that serves British cuisine in the city center, with the reservation set for one person on a Wednesday at 12:30 PM. Additionally, the user requires transportation arrangements to be made from a train station to the restaurant, ensuring arrival by the reserved time. The user is looking for recommendations on suitable dining establishments that fit these criteria and for a taxi service that can accommodate the specified departure and arrival details.\\
		\bottomrule
	\end{tabular}
	\caption{An example of user goal generation. This prompt template is further improved by ChatGPT after we manually construct it.}
	\label{prompt-template-user-goal}
\end{table*}
\begin{table*}
	\centering
	\begin{tabular}{p{2\columnwidth}}
		\toprule
		\textbf{Prompt:}\\
		\#\# Dialogue State: <synthetic dialogue state>\\
		\#\# User Goal: <generated user goal>\\
		\#\# Information for the Agent: <additional information in the database for a certain place>\\
		\#\# Instruction:\\
		The dialogue state summarize the dialogue between the user and the agent. Your task is to generate a dialogue flow. Each element in the dialogue flow determine the information of a dialogue turn. Each element should contain a description for the dialogue turn and the corresponding turn state. The description is used to guide the subsequent turn utterance generation. The state is a subset of the dialogue state that describe the information the user mentioned or confirmed in the dialogue turn. Note that the dialogue is started with the agent and ended with the user and must follow an alternating pattern between the agent and the user. After a successful booking service, the model should provide a synthetic reference number, which consists of 8 random characters. Sometimes there are a lot of restriction for a domain, so you should construct more dialogue turns to make sure that the user express no more than six new restrictions of domain-slots in one dialogue turn. The value `dontcare' means that the user does not care about the restriction of that domain-slot and this information must be expressed by the user.
		
		Remember that all the states in the dialogue state should be used and no additional information can be added.
		
		Remember that the agent does not have prior knowledge of the user's goals.
		
		Ensure that the user express no more than six new restriction of domain-slots in one dialogue turn. Be meticulous to confirm that the user does not introduce more than six new restrictions of domain-slots within a single turn. To comply with this, divide constraints across multiple turns when necessary.\\
		
		\midrule
		\textbf{Parsed GPT output:} \\
		\{
			"description": "The agent greets the user and asks what they can assist them with today.",\\
			"turn state": \{\}
		\},\\
		\{
			"description": "The user requests to book a table at a high-end restaurant that serves British cuisine in the city center for one person.",\\
			"turn state":\{
				"restaurant-area": "centre",
				"restaurant-food": "british",
				"restaurant-book people": "1",
				"restaurant-pricerange": "expensive"
			\}
		\},\\
		... ,\\
		\{
			"description": "The user confirms the booking at Midsummer House Restaurant and requests a taxi to get there, specifying they need to arrive by 12:30 PM.",\\
			"turn state": \{
				"taxi-destination": "midsummer house restaurant",
				"taxi-arriveby": "12:30"
			\}
		\},... \\
		\bottomrule
	\end{tabular}
	\caption{An example of dialogue flow generation.}
	\label{prompt-template-dialogue-flow}
\end{table*}

\begin{table*}
	\centering
	\begin{tabular}{p{2\columnwidth}}
		\toprule
		\textbf{Prompt:}\\
		\#\# Information for the Agent: <additional information in the database for a certain place>\\
		\#\# Dialogue Flow: <generated dialogue flow>\\
		\#\# Instruction:\\
		Based on the dialogue flow, please generate a dialogue between a agent and a user. You should include information provided in the description and the turn state while construct a dialogue turn. The synthetic reference number should be a string consists of 8 random character. Remember no turn in the dialogue flow can be omitted and no additional dialogue turn can be added. The generated dialogue turns must correspond one-to-one with the provided dialogue flow of, neither more nor less.
		
		Remember no turns in the dialogue flow can be omitted and no additional dialogue turn can be added.
		
		\#\# Output Format: <output format>\\
		\midrule
		\textbf{Parsed GPT output:}\\
		\{\\
		"role": "agent",\\
		"description": "The agent greets the user and asks what they can assist them with today.",\\
		"content": "Good day! How may I assist you today?"\\
		\},
		
		\{\\
		"role": "user",\\
		"description": "The user requests to book a table at a high-end restaurant that serves British cuisine in the city center for one person.",\\
		"content": "I'd like to book a table for one at a high-end British cuisine restaurant in the city center."\\
		\},
		
		...\\
		\{\\
		"role": "agent",\\
		"description": "The agent confirms the reservation details and informs the user that they have a table available at Midsummer House Restaurant, which fits all the criteria provided.",\\
		"content": "I have a table available for you at Midsummer House Restaurant, which offers British cuisine and is located in the city center. It's quite an exquisite place, and it fits your criteria perfectly."\\
		\},
		
		\{\\
		"role": "user",\\
		"description": "The user confirms the booking at Midsummer House Restaurant and requests a taxi to get there, specifying they need to arrive by 12:30 PM.",\\
		"content": "That sounds wonderful. Please confirm my booking at Midsummer House Restaurant. Also, I need a taxi to get there and arrive by 12:30 PM."\\
		\}\\
		...\\
		\bottomrule
	\end{tabular}
	\caption{An example of dialogue generation.}
	\label{prompt-template-dialogue-generation}
\end{table*}
\begin{table*}
	\centering
	\begin{tabular}{p{2\columnwidth}}
		\toprule
		\textbf{Prompt:}\\
		\#\# Dialogue History: <dialogue history>\\
		\#\# Current Turn Utterances:\\
		system: i have a table available for you at midsummer house restaurant , which offers british cuisine and is located in the city center . it s quite an exquisite place , and it fits your criteria perfectly.\\
		user: that sounds wonderful . please confirm my booking at midsummer house restaurant . also , i need a taxi to get there and {\color[RGB]{255, 0, 0}{arrive by 12:30 pm}}.\\
		\#\# Turn State: <turn state>\\
		\#\# Co-reference:\\
		\{"taxi-destination": "restaurant-name",
		"taxi-arriveby": "restaurant-book time"\} \\
		\#\# Instruction:\\
		Your task is to modify the current turn utterances to express the co-reference information implicitly rather than explicitly stating the full name of the value. Note that the rest of the values in the current turn state should still be expressed explicitly. The modified utterance should be the paraphrase of the original utterance in the current turn for both the system and the user. Only to modify the information for these co-reference value.
		
		Remember to maintain the meaning of the modified utterance with the original one. 
		
		If the expression of the co-reference value in the original utterance is already in a implicit way, please do not modify the utterances and just copy it.
		
		\#\# Output Format:\\
		\{\\
			“description”: <description of the co-reference information in natural language>,\\
			“system”: <modified system utterance of the current turn>,\\
			“user”: <modified user utterance of the current turn>\\
		\}\\
		\midrule
		\textbf{Parsed GPT output:} \\
		\{\\
		"description": "The `taxi-destination' is the same as the `restaurant-name', which is Midsummer House Restaurant, and the `taxi-arriveby' time is the same as the `restaurant-book time', which is 12:30 pm.", \\
		"system": "I have a table available for you at Midsummer House Restaurant, which offers British cuisine and is located in the city center. It's quite an exquisite place, and it fits your criteria perfectly.", \\
		"user": \"That sounds wonderful. Please confirm my booking at Midsummer House Restaurant. Also, I need a taxi to get there {\color[RGB]{84, 180, 53}{in time for my reservation}}.\\
		\} \\
		\bottomrule
	\end{tabular}
	\caption{An example of dialogue complication. As described in "\# Co-reference", taxi-arriveby should be inferred from restaurant-book time. GPT modify the expression of that information. Words in red are the expression before modification and words in green are the expression after modification.}
	\label{prompt-template-dialogue-complication}
\end{table*}
\begin{table*}
	\centering
	\begin{tabular}{p{2\columnwidth}}
		\toprule
		\textbf{The original training sample:}\\
		\#\# Dialogue History: <dialogue history>\\
		\#\# Current Turn Utterances:\\
		system: What can I help you today?
		user: I want to have dinner in a restaurant that serves Chinese food, and it should be in the city centre.\\
		\#\# Target Output: "Chinese | centre"\\
		\midrule
		\textbf{After permutation:} \\
		\textbf{Training sample-1:} \\
		\#\# Dialogue History: <dialogue history>\\
		\#\# Current Turn Utterances:\\
		system: What can I help you today?
		user: I want to have dinner in a restaurant that serves Chinese food, and it should be in the city centre.\\
		\#\# Target Output: {\color[RGB]{84, 180, 53}{"Chinese | centre"}}\\
		\textbf{Training sample-2:} \\
		\#\# Dialogue History: <dialogue history>\\
		\#\# Current Turn Utterances:\\
		system: What can I help you today?
		user: I want to have dinner in a restaurant that serves Chinese food, and it should be in the city centre.\\
		\#\# Target Output: {\color[RGB]{84, 180, 53}{"centre | Chinese"}}\\
		\bottomrule
	\end{tabular}
	\caption{An example of slot value permutation. For each training sample in the dataset, we will permute the slot values. Therefore, the sample containing 2 values will be augmented to 2 samples, the sample containing 3 values will be augmented to 6 samples. And so on, for each sample.}
	\label{permutation-example}
\end{table*}
\end{document}